\RequirePackage{fix-cm}
\documentclass[twocolumn]{svjour3}          
\smartqed  
\usepackage{graphicx}
\usepackage{natbib}

\usepackage{soul}
\usepackage{url}
\usepackage[utf8]{inputenc}
\usepackage[small]{caption}
\usepackage{subcaption}
\usepackage{graphicx}
\usepackage{amsmath}
\usepackage{booktabs}
\usepackage{algorithm}
\usepackage{algorithmic}
\usepackage{epsfig}
\usepackage{amssymb}
\usepackage{color}
\usepackage[table]{xcolor}
\usepackage{hhline}
\usepackage{stfloats}
\definecolor{Gray}{gray}{0.85}
\usepackage{diagbox}
\usepackage{multirow}
\usepackage{comment}

\newcommand{\eqnref}[1]{Eq.~(\ref{eqn:#1})}
\newcommand{\figref}[1]{Fig.~\ref{fig:#1}}
\newcommand{\tabref}[1]{Table~\ref{tbl:#1}}

\usepackage[pagebackref=true,breaklinks=true,colorlinks=True,bookmarks=false,linkcolor=blue,citecolor=blue]{hyperref}

\usepackage{dsfont}
\usepackage{pifont} 

\newcommand{\xmark}{\ding{55}}%
\usepackage[pagebackref=true,breaklinks=true,colorlinks=True,bookmarks=false,linkcolor=blue,citecolor=blue]{hyperref}

\newcommand{\synzebra}{
\begin{figure*}[t]
  \centering
  \includegraphics[width=0.95\linewidth]{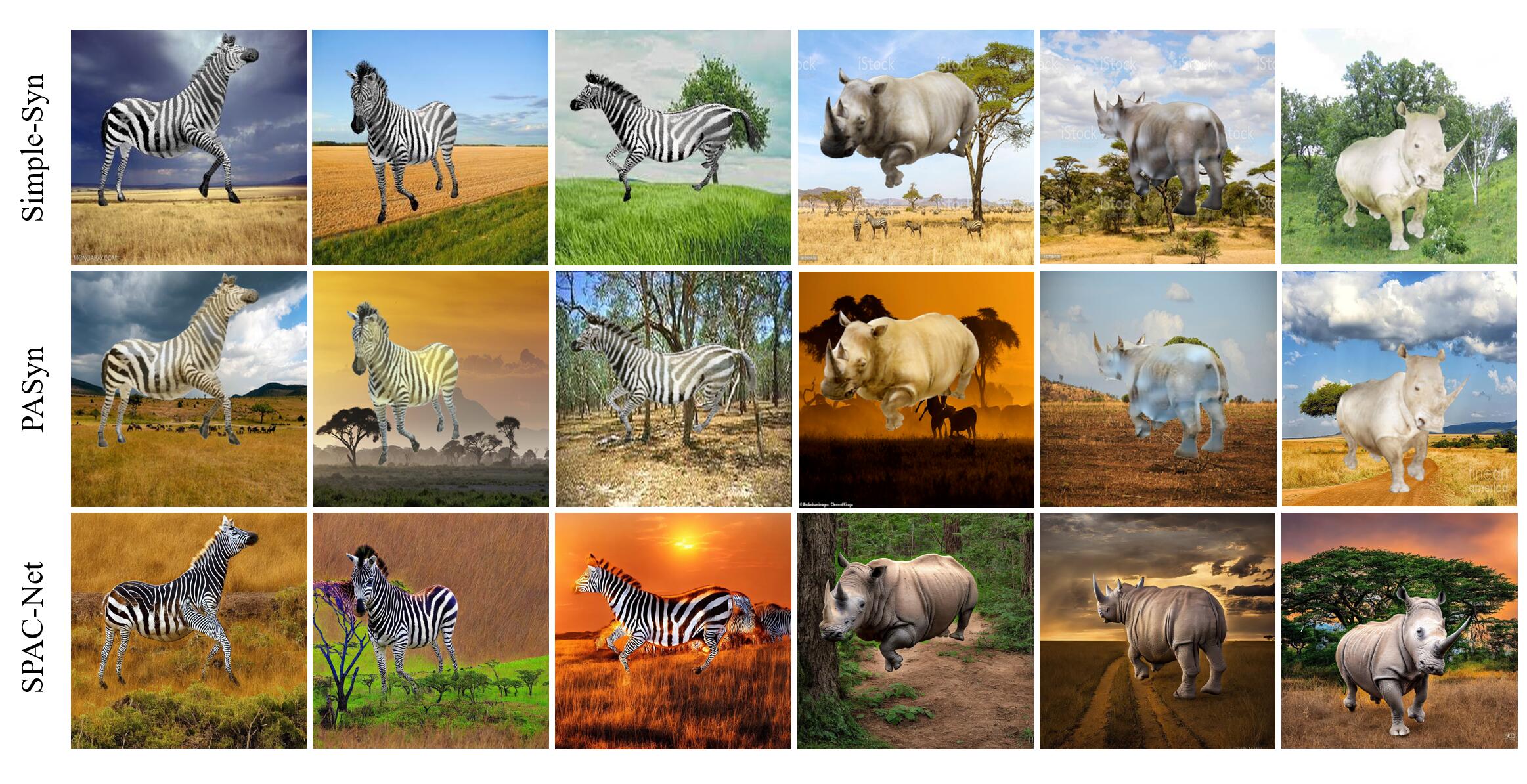}
  \caption{Comparison of synthetic animal images generated by three different methods: (1) Simple Synthetic (Simple-Syn), where animals are directly placed on background images without style transfer, (2) PASyn \cite{jiang2022prior}, which is used a transformer-based style transfer approach, StyTr$^2$ \cite{deng2022stytr2}, to transfer the background style onto the animal, incorporating environmental lighting effects on the texture, and (3) SPAC-Net, the proposed method in this work, which uses the Holistically-nested Edge Detection (HED) boundary task in ControlNet for style transfer from the synthetic domain to the real domain.}
  \label{fig:synzebra}
\end{figure*}
}

\newcommand{\PASyC}{
\begin{figure*}[t]
  \centering
  \includegraphics[width=1\linewidth]{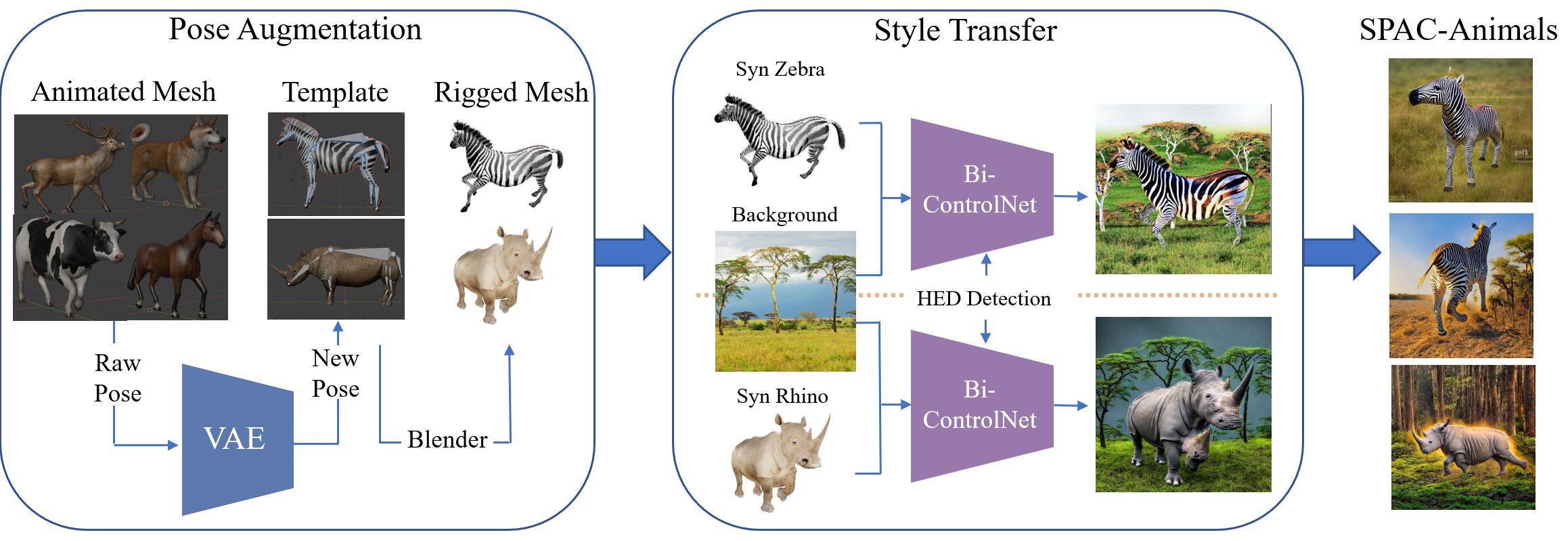}
  \caption{An overview architecture of our synthetic prior-aware animal ControlNet (SPAC-Net), composed of three parts: pose augmentation, style transfer and dataset generation. The SPAC-Net pipeline leads to generation of our probabilistically-valid animal SPAC-Animals dataset.}
  \label{fig:PASyC}
\end{figure*}
}

\newcommand{\armature}{
\begin{figure}[t]
  \centering
  \includegraphics[width=0.95\linewidth]{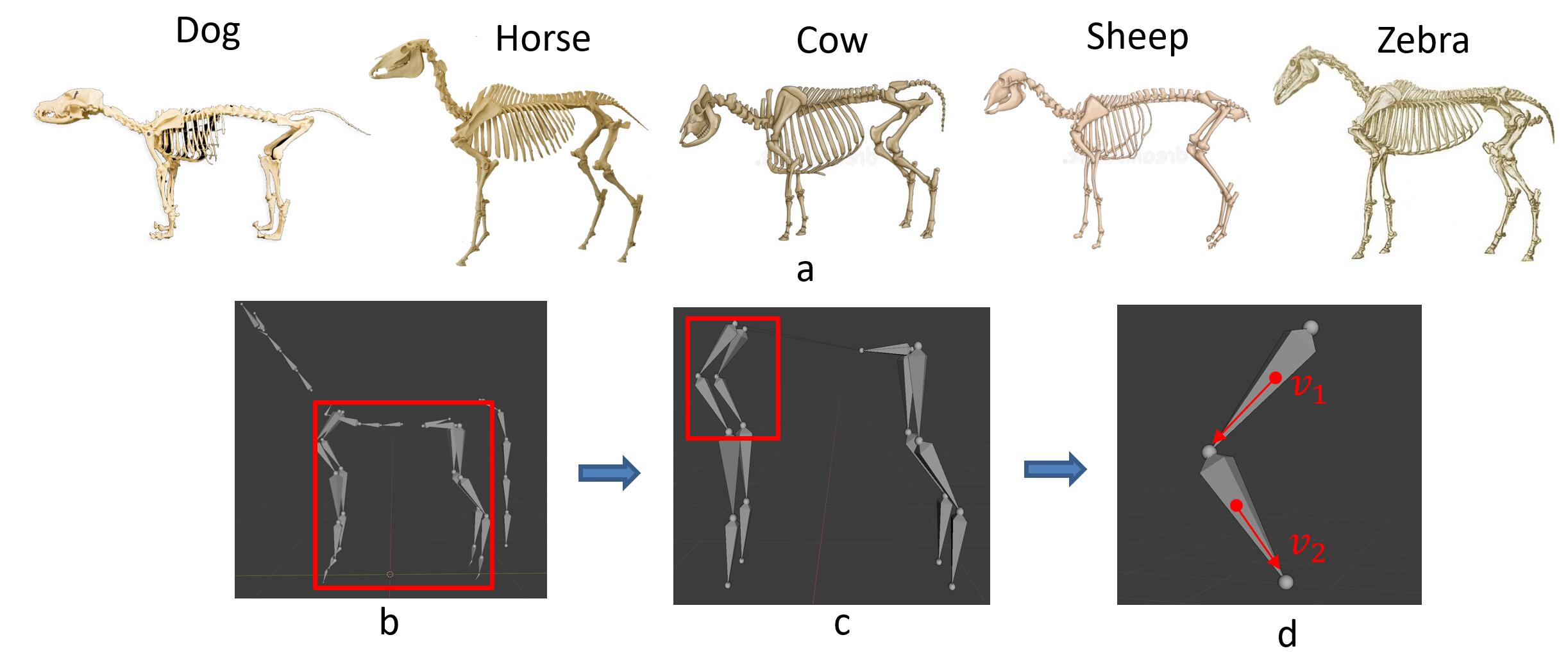}
  \caption{General quadruped armature. (a) shows the real skeletons of horse (bottom) and dog (top) respectively. (b) shows the artificially designed animal skeleton based on the real one. The red box marks the skeleton of limbs which we are interested in. (c) shows the skeleton of limbs. In (d), v1 and v2 represent the spatial orientation of two adjacent bones. The angle between the vectors is the angle between the two bones.}
  \label{fig:armature}
\end{figure}
}

\newcommand{\VAE}{
\begin{figure*}[t]
  \centering
  \includegraphics[width=1.0\linewidth]{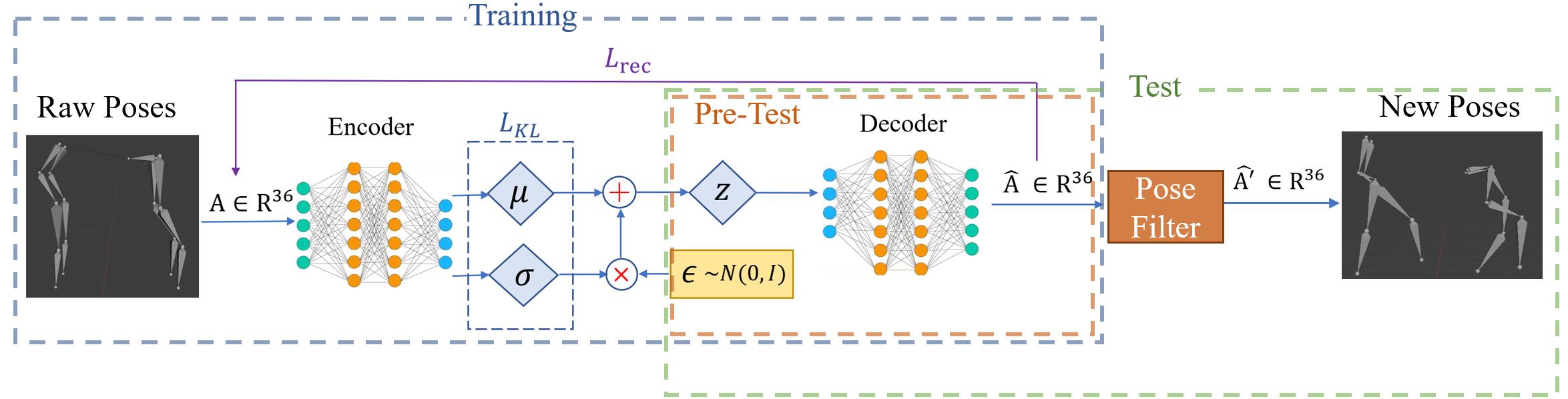}
  \caption{An overview architecture of our VAE-based animal poses generative model, composed of two main parts: training (blue-dot box) and test (green-dot box). Before we run “test” to generate new poses, pre-test (brown-dot box) should be performed independently to specify the sampling value ranges for each angle in the ‘pose filter.}
  \label{fig:VAE}
\end{figure*}
}

\newcommand{\issuecontrolnet}{
\begin{figure}[ht]
  \centering
  \includegraphics[width=1.0\linewidth]{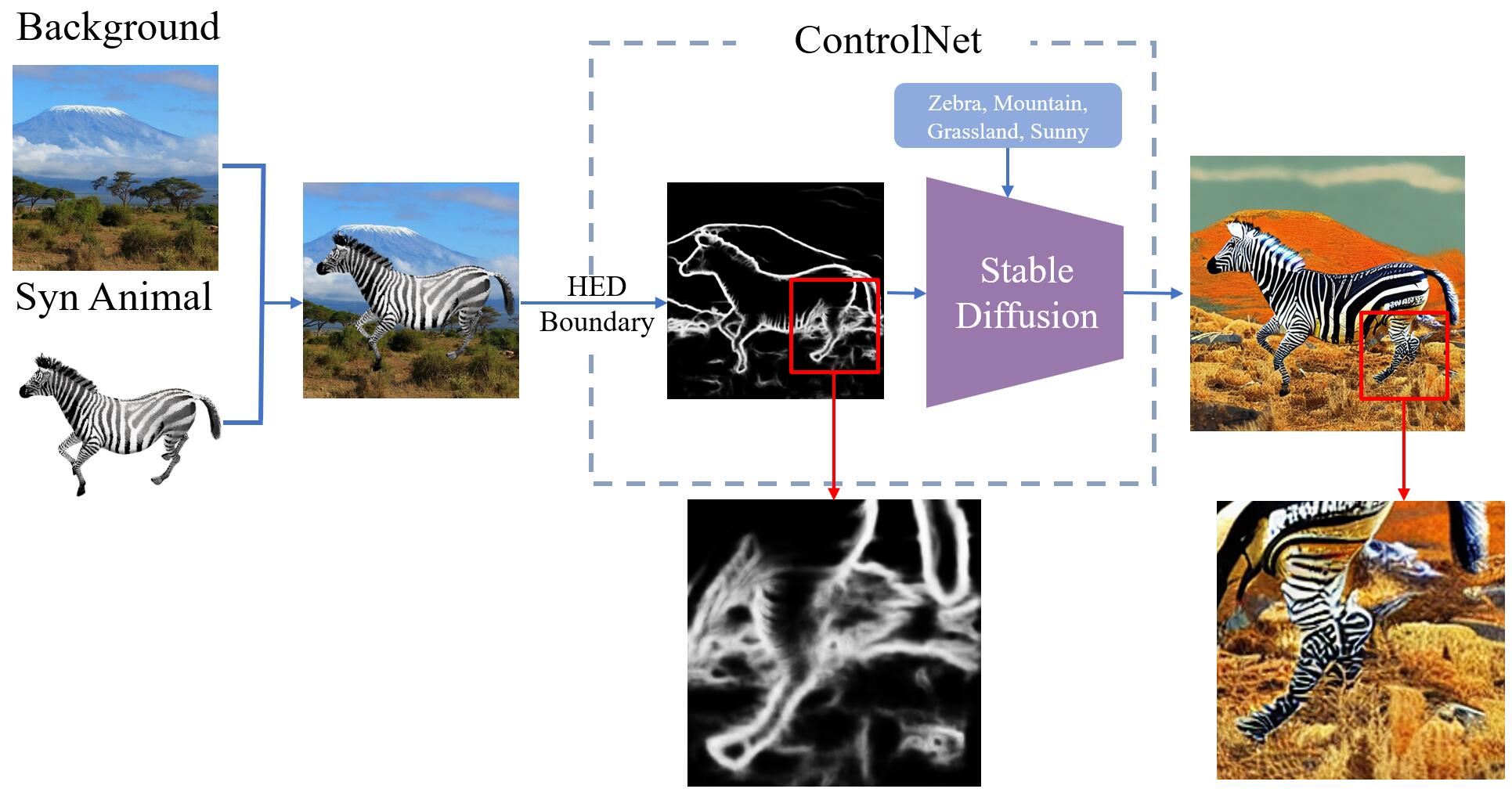}
  \caption{The issue with direct ControlNet use. In the figure, we use red boxes to highlight the areas where limb blur appears.}
  \label{fig:issuecontrolnet}
\end{figure}
}
\newcommand{\bicontrolnet}{
\begin{figure}[h]
  \centering
  \includegraphics[width=1.0\linewidth]{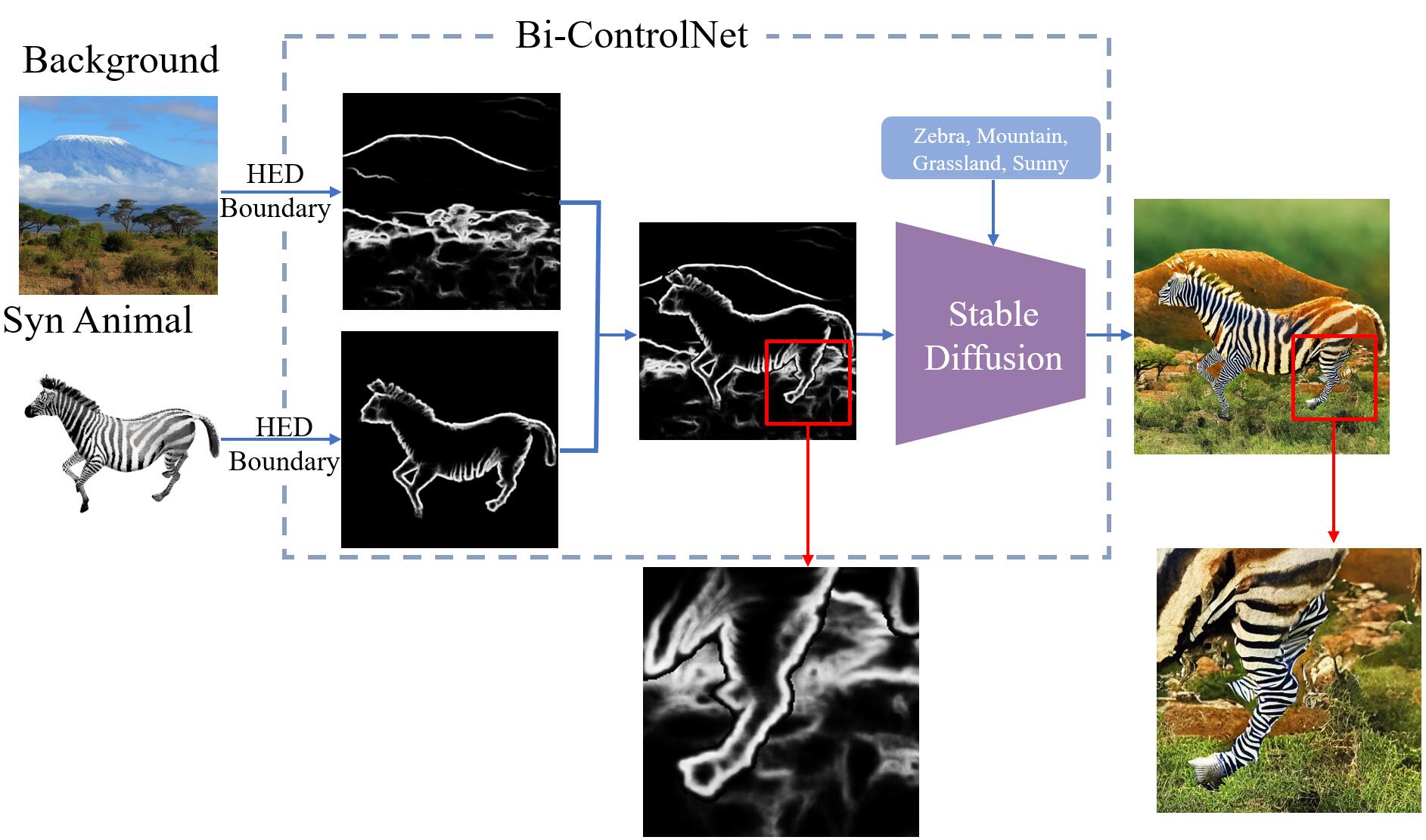}
  \caption{The Bi-ControlNet architecture, which separates the detection of the HED boundary for the background and the subject. This approach helps to minimize the likelihood of Stable Diffusion misidentifying the location of the zebra.}
  \label{fig:bicontrolnet}
\end{figure}
}

\newcommand{\posefilter}{
\begin{figure}[h]
    \centering
    \includegraphics[width=0.99\linewidth]{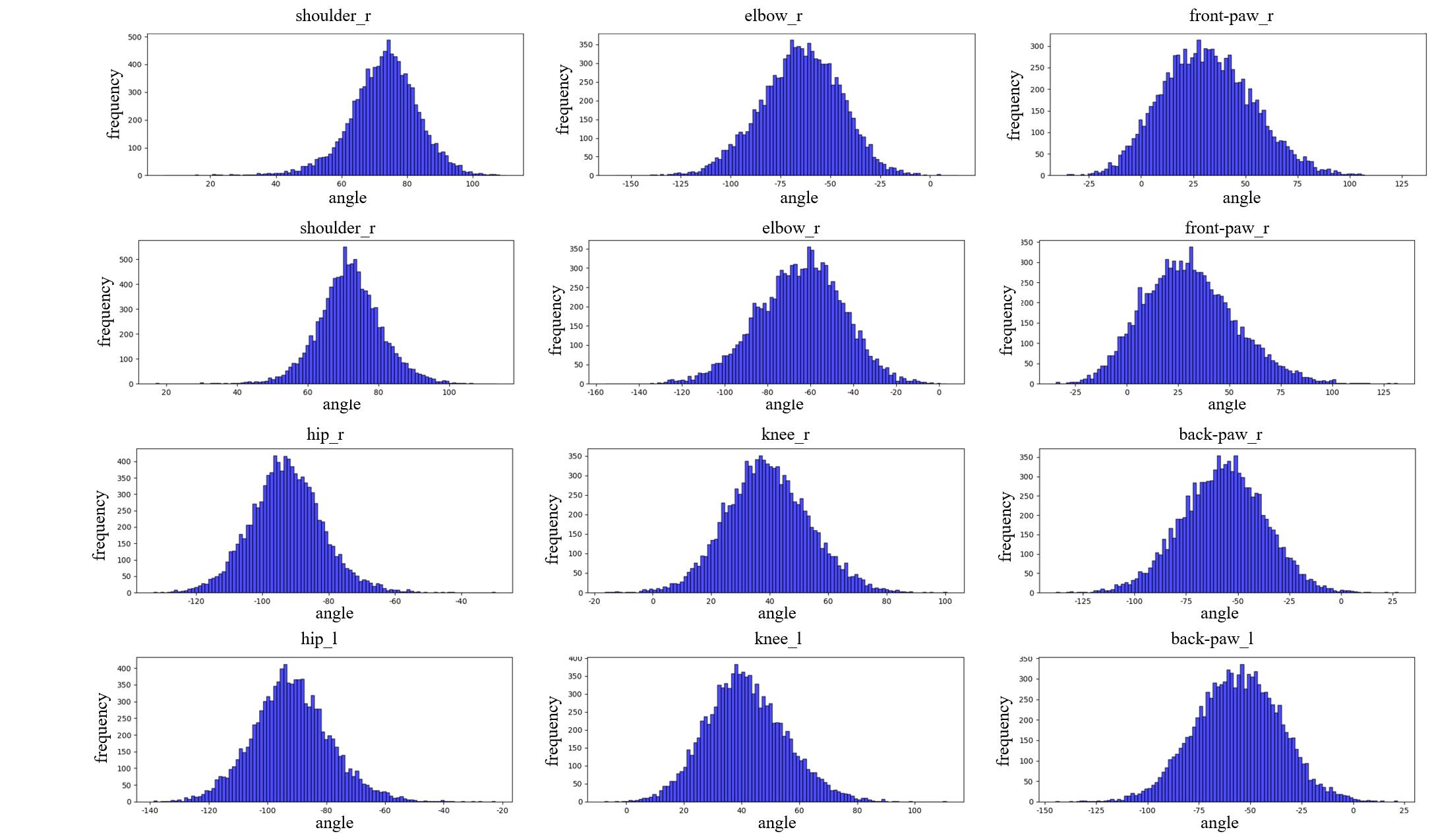}
    \caption{The histograms of angle values of 12 joints, which are shoulder\_right, elbow\_right, front-paw\_right, shoulder\_left, elbow\_left , front-paw\_left, hip\_right, knee\_right, back-paw\_right, hip\_left, knee\_left , back-paw\_left. }
    \label{fig:posefilter}
            \vspace{-.2in}
\end{figure}
}

\newcommand{\ablation}{
\begin{figure*}[t]
  \centering
  \includegraphics[width=0.95\linewidth]{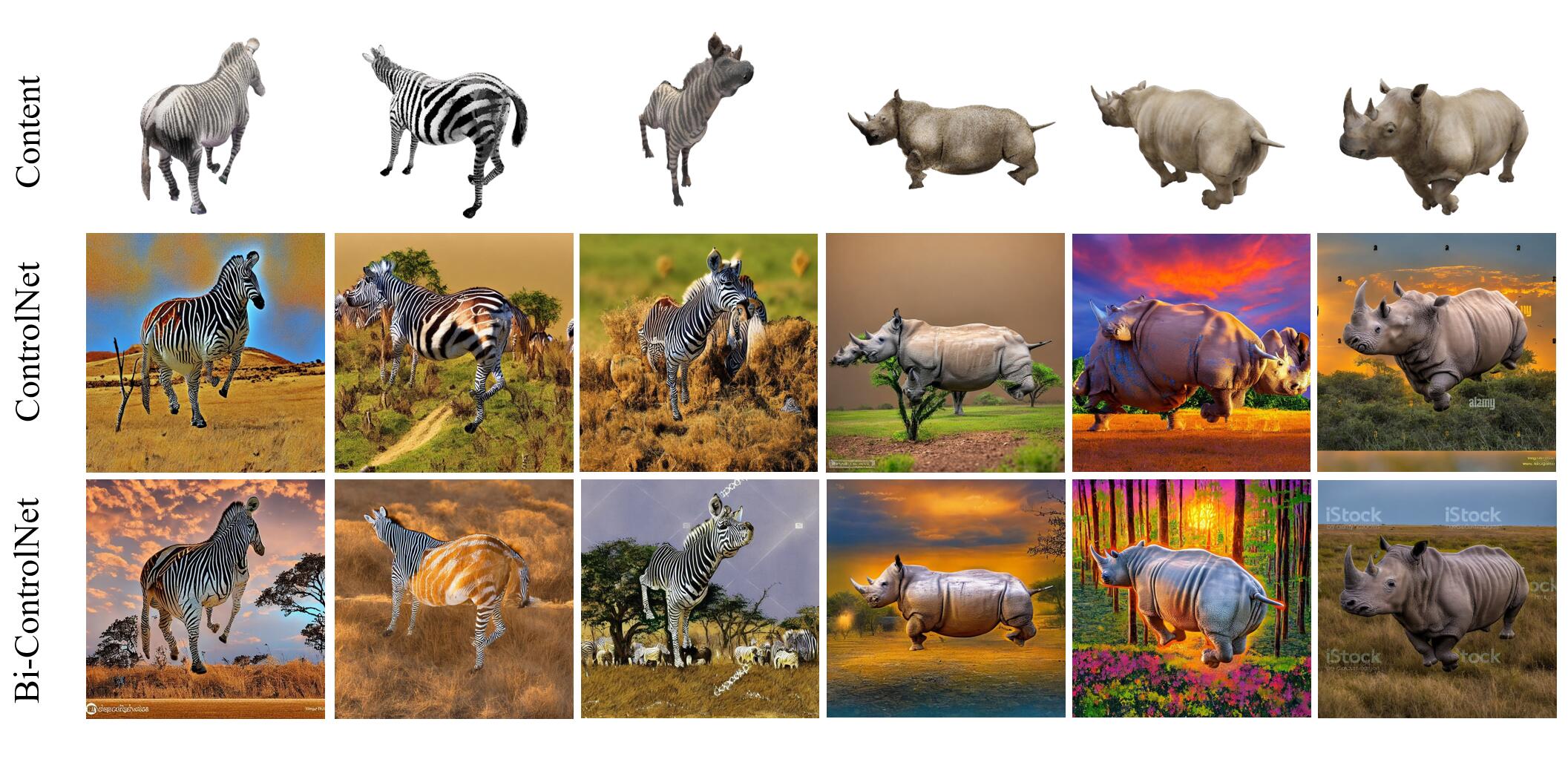}
  \caption{Ablation Study: The first row represents the synthetic animals used as pose references. The second and third rows display the synthetic images generated by the standard ControlNet and Bi-ControlNet, respectively.}
  \label{fig:ablation}
\end{figure*}
}

\newcommand{\tsne}{
\begin{figure}[b]
  \centering
  \includegraphics[width=0.99\linewidth]{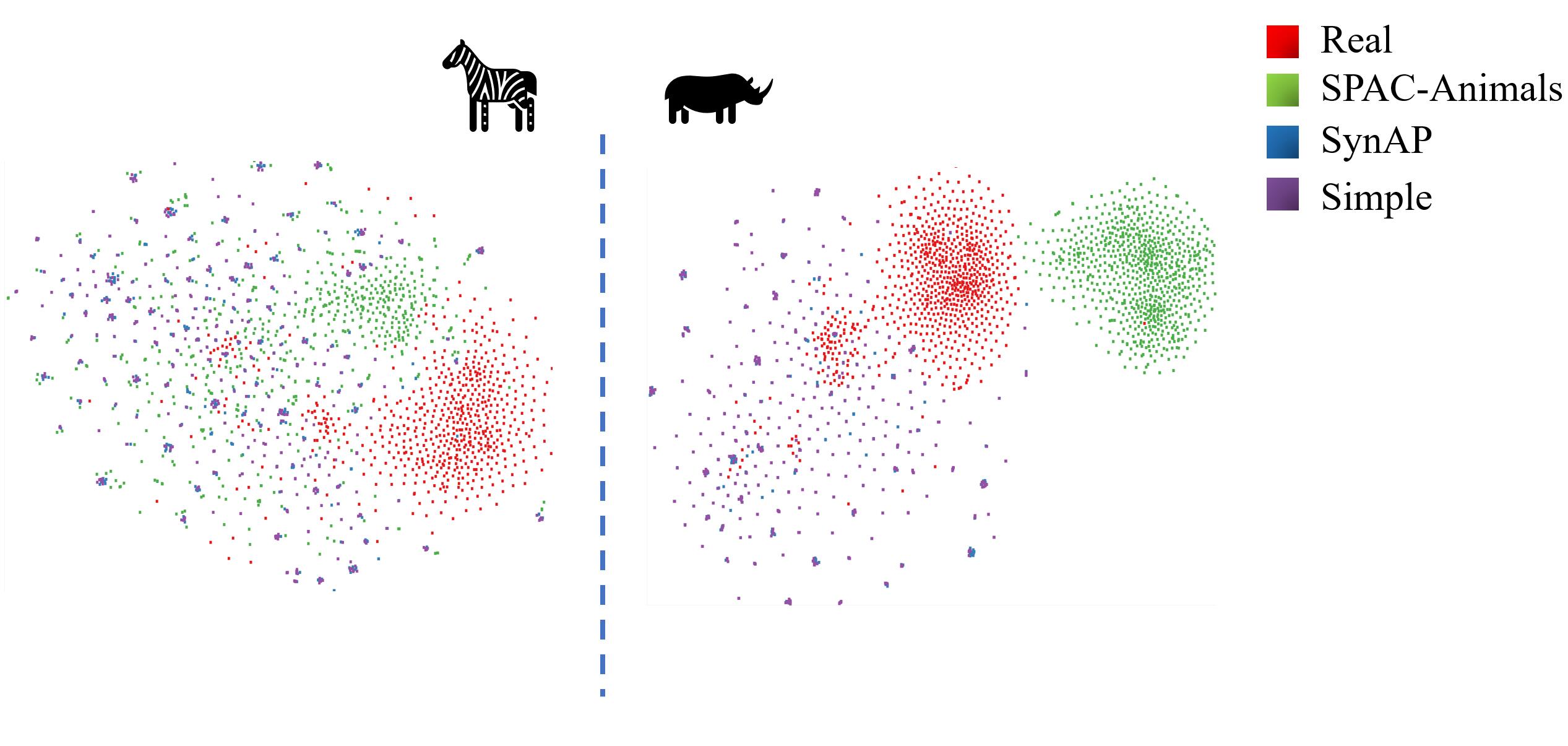}
  \caption{This figure presents a visual comparison of data from four distinct domains: real (in red), simple synthetic (in purple), SynAP (in blue), and SPAC-animals (in green). The t-SNE visualization technique is applied using the output of the HRNet-w32 at stage 2 as the feature representation. On the left, the results for zebras are displayed, while on the right, the results for rhinos are showcased. It is important to note that the feature extractors were trained independently on 99 real images for both zebras and rhinos.}
  \label{fig:tsne}
\end{figure}
}
\usepackage{multirow}
\newcommand{\mainresult}{
\begin{table*}[t]
\caption{The effect of SPAC-Animals with limited real data of pose estimation results of the common backbones, HRNet-w32 \cite{hrnet} tested on zebra-300 and rhino-300 (300 real images for each test set). The training set contains only 99 real zebra or rhino images (from AP10K)  and augmented with SynAP or SPAC-Animals. Each of them includes 3,000 synthetic zebra and rhino images. Best results for each backbone are shown in bold.}
\vspace{-.4cm}
\begin{center}
\resizebox{\textwidth}{!}{
\begin{tabular}{l l c c c c c c c c c c c c}
\multicolumn{1}{c}{\multirow{2}{*}{Method}}     & \multicolumn{1}{c}{\multirow{2}{*}{Training Set}} & \multicolumn{1}{l|}{\multirow{2}{*}{Test Animal}} & \multicolumn{11}{c}{PCK@0.05 Pose Estimation Accuracy on Zebra-300 and Rhino-300 Set}   \\ \cline{4-14} 
\multicolumn{1}{c}{}                            & \multicolumn{1}{c}{}                              & \multicolumn{1}{l|}{}                             & Eye & Nose & Neck & Shoulders & Elbows & F-Paws & Hips & Kness & B-Paws & RoT & Average \\ \hline
\multicolumn{1}{l|}{\multirow{4}{*}{MMPose}} & \multicolumn{1}{l|}{R(99)}                        & \multicolumn{1}{c|}{\multirow{4}{*}{Zebra}}       & 97.3&95.8 & 83.2 &78.8& 77.1 &62.6  &86.0 &74.9 &59.8 & 82.4&78.7  \\ \cline{2-2} \cline{4-14} 
\multicolumn{1}{l|}{}                           & \multicolumn{1}{l|}{R(99)+AP10K(8K)}              & \multicolumn{1}{c|}{}                          & 97.5&97.2 & 79.4  & 87.8 & 90.3 & \textbf{93.8}& 95.3 & 94.1 & 89.5 &86.4 &91.4 \\ \cline{2-2} \cline{4-14} 
\multicolumn{1}{l|}{}                           & \multicolumn{1}{l|}{R(99)+SynAP(3K)}              & \multicolumn{1}{c|}{}                          & \textbf{97.8}&\textbf{98.3}&81.1&94.0&93.5&92.0&93.7&93.5&89.0&87.6&92.4    \\ \cline{2-2} \cline{4-14} 
\multicolumn{1}{l|}{}                           & \multicolumn{1}{l|}{R(99)+SPAC(3K)}               & \multicolumn{1}{c|}{}                          & \textbf{97.8}& 96.5&\textbf{93.4}&\textbf{98.4}&\textbf{95.5} & 92.9&\textbf{98.2} &\textbf{96.9}& \textbf{95.7}& \textbf{97.2}& \textbf{96.3}        \\ \hline \hline
\multicolumn{1}{l|}{\multirow{4}{*}{MMPose}} & \multicolumn{1}{l|}{R(99)}                        & \multicolumn{1}{c|}{\multirow{4}{*}{Rhino}}       & 93.1&\textbf{99.7}&77.0&93.8&91.0&86.5&84.2&92.9&72.3&97.0& 88.3 \\\cline{2-2} \cline{4-14} 
\multicolumn{1}{l|}{}                           & \multicolumn{1}{l|}{R(99)+AP10K(8K)}              & \multicolumn{1}{c|}{}                          & 98.1&98.6&82.4&\textbf{98.6}&\textbf{97.8}&\textbf{97.9}&93.5&\textbf{98.5}&\textbf{98.6}&\textbf{98.5}&\textbf{96.7}\\ \cline{2-2}\cline{4-14} 
\multicolumn{1}{l|}{}                           & \multicolumn{1}{l|}{R(99)+SynAP(3K)}              & \multicolumn{1}{c|}{}                          &99.6&\textbf{99.7}&\textbf{83.4}&98.4&97.3&96.4&93.7&96.2&94.5&97.7&95.9\\\cline{2-2} \cline{4-14} 
\multicolumn{1}{l|}{}                           & \multicolumn{1}{l|}{R(99)+SPAC(3K)}               & \multicolumn{1}{c|}{}                          &\textbf{99.7}&98.0&81.8&97.5&96.4&97.1&\textbf{94.2}&96.5&96.2&\textbf{98.5}&95.9\\ \hline
                                                &                                                   & \multicolumn{1}{l}{}                              &     &      &      &           &        &        &      &       &        &     &         \\
                                                &                                                   & \multicolumn{1}{l}{}                              &     &      &      &           &        &        &      &       &        &     &         \\
                                                &                                                   & \multicolumn{1}{l}{}                              &     &      &      &           &        &        &      &       &        &     &         \\
                                                &                                                   & \multicolumn{1}{l}{}                              &     &      &      &           &        &        &      &       &        &     &         \\
                                                &                                                   & \multicolumn{1}{l}{}                              &     &      &      &           &        &        &      &       &        &     &        
\end{tabular} }
\vspace{-1cm}
\label{tbl:mainresult}
\end{center}
\end{table*}
}

\newcommand{\abl}{
\begin{table*}[t]
\caption{Ablation study on two parts of  SPAC-Net pipeline, VAE , style transfer on two different zebra datasets with resolution of 300$\times$300. The HRNet-w32 is the backbone here. The training set contains 99 real zebra (from AP10K) and augmented with SPAC-Animals (3,000 synthetic zebra images). Best results are shown in bold.}

\vspace{-.3cm}
\begin{center}
\begin{tabular}{c|c|c|c|c|c}
\footnotesize
\begin{tabular}[c]{@{}l@{}}\textbf{Index}\end{tabular}&
\begin{tabular}[c]{@{}l@{}}\textbf{Training Set}\end{tabular} & 
\begin{tabular}[c]{@{}l@{}}\textbf{VAE}\end{tabular} & 
\begin{tabular}[c]{@{}l@{}}\textbf{Style Transfer}\end{tabular} & 

\begin{tabular}[c]{@{}l@{}}\textbf{Zoo Zebra Set}\end{tabular} & 
\begin{tabular}[c]{@{}l@{}}\textbf{Zebra-300 Set} \end{tabular}  \\ \hline
a  & R(99) & \xmark & \xmark&  76.0 & 78.7\\ \hline \hline
b  & S(3K) & \xmark & \xmark &38.7 & 30.0 \\ \hline
c  & S(3K) & \checkmark  & \xmark &44.2& 36.7\\ \hline 
d  & S(3K) & \checkmark  & \checkmark &73.2 & 80.4\\ \hline \hline

e  & R(99)+S(3K) & \xmark & \xmark&89.8&88.0 \\ \hline
f  & R(99)+S(3K) & \checkmark  & \xmark  &90.4&89.8 \\ \hline
g  & R(99)+S(3K) & \checkmark  & \checkmark  & \textbf{94.9}&\textbf{96.2} \\ \hline

\end{tabular}
\vspace{-.5cm}
\label{tbl:abl}
\end{center}
\end{table*}
}
\newcommand{\syncompare}{
\begin{table*}[t]
\caption{A comparative analysis of different types of synthetic data, including Simple Synthetic, SynAP, ControlNet (directly feeding Simple Synthetic data into ControlNet), and Bi-ControlNet (generated by SPAC-Net) is conducted. Each type of synthetic data comprises 3,000 animal images. All experiments employ HRNet-w32 as the backbone and are tested on the Zebra-300 and Rhino-300 test sets. Best results for each backbone are shown in bold.}
\vspace{-.4cm}
\begin{center}
\resizebox{\textwidth}{!}{
\begin{tabular}{l l c c c c c c c c c c c c}
\multicolumn{1}{c}{\multirow{2}{*}{Method}}     & \multicolumn{1}{c}{\multirow{2}{*}{Training Set}} & \multicolumn{1}{l|}{\multirow{2}{*}{Test Animal}} & \multicolumn{11}{c}{PCK@0.05 Pose Estimation Accuracy on Zebra-300 and Rhino-300 Set}   \\ \cline{4-14} 
\multicolumn{1}{c}{}                            & \multicolumn{1}{c}{}                              & \multicolumn{1}{l|}{}                             & Eye & Nose & Neck & Shoulders & Elbows & F-Paws & Hips & Kness & B-Paws & RoT & Average \\ \hline
\multicolumn{1}{l|}{\multirow{5}{*}{MMPose}} & \multicolumn{1}{l|}{R(99)}                        & \multicolumn{1}{c|}{\multirow{5}{*}{Zebra}}      &\textbf{97.3}&\textbf{95.8} & \textbf{83.2} &78.8& 77.1 &\textbf{62.6}  &86.0 &74.9 &59.8 & 82.4&78.7\\ \cline{2-2} \cline{4-14} 
\multicolumn{1}{l|}{}                           & \multicolumn{1}{l|}{Simple(3K)}                   & \multicolumn{1}{c|}{}                         &30.7&19.9&31.1&48.0&34.1&36.4&41.9&38.3&34.0&45.6&36.7\\ \cline{2-2} \cline{4-14} 
\multicolumn{1}{l|}{}                           & \multicolumn{1}{l|}{SynAP(3K)}                    & \multicolumn{1}{c|}{}                         &47.1&39.9&36.0&64.7&38.9&27.9&55.4&52.9&38.0&61.6&46.6\\ \cline{2-2} \cline{4-14} 
\multicolumn{1}{l|}{}                           & \multicolumn{1}{l|}{ControlNet(3K)}               & \multicolumn{1}{c|}{}                 &88.8&81.8&53.1&78.2&62.8&58.1&67.6&76.6&\textbf{73.0}&66.4&70.9\\ \cline{2-2} \cline{4-14} 
\multicolumn{1}{l|}{}                           & \multicolumn{1}{l|}{Bi-ControlNet(3K)}            & \multicolumn{1}{c|}{}                 &84.7&73.4&78.3&\textbf{89.4}&\textbf{78.1}&\textbf{62.6}&\textbf{91.4}&\textbf{83.5}&68.4&\textbf{94.8}&\textbf{80.4}\\ \hline \hline
\multicolumn{1}{l|}{\multirow{5}{*}{MMPose}} & \multicolumn{1}{l|}{R(99)}                        & \multicolumn{1}{c|}{\multirow{5}{*}{Rhino}}      & \textbf{93.1}&\textbf{99.7}&\textbf{77.0}&\textbf{93.8}&\textbf{91.0}&\textbf{86.5}&84.2&\textbf{92.9}&\textbf{72.3}&\textbf{97.0}&\textbf{ 88.3}         \\ \cline{2-2} \cline{4-14} 
\multicolumn{1}{l|}{}                           & \multicolumn{1}{l|}{Simple(3K)}                   & \multicolumn{1}{c|}{}                             &33.2&28.4&30.1&28.4&18.0&14.7&50&38.1&22.8&47.0&29.9\\ \cline{2-2} \cline{4-14} 
\multicolumn{1}{l|}{}                           & \multicolumn{1}{l|}{SynAP(3K)}                    & \multicolumn{1}{c|}{}                             &83.3&78.7&68.6&71.9&55.8&40.3&83.8&70.2&35.9&87.2&64.9\\ \cline{2-2} \cline{4-14} 
\multicolumn{1}{l|}{}                           & \multicolumn{1}{l|}{ControlNet(3K)}               & \multicolumn{1}{c|}{}                             &68.3&67.3&66.6&72.4&59.4&51.1&81.2&69.2&51.7&81.2&65.8\\ \cline{2-2} \cline{4-14} 
\multicolumn{1}{l|}{}                           & \multicolumn{1}{l|}{Bi-ControlNet(3K)}            & \multicolumn{1}{c|}{}                             &80.6&77.3&72.0&85.0&67.1&46.0&\textbf{91.1}&77.3&46.3&86.8&71.5\\ \hline
                                                &                                                   & \multicolumn{1}{l}{}                              &     &      &      &           &        &        &      &       &        &     &         \\
                                                &                                                   & \multicolumn{1}{l}{}                              &     &      &      &           &        &        &      &       &        &     &         \\
                                                &                                                   & \multicolumn{1}{l}{}                              &     &      &      &           &        &        &      &       &        &     &         \\
                                                &                                                   & \multicolumn{1}{l}{}                              &     &      &      &           &        &        &      &       &        &     &         \\
                                                &                                                   & \multicolumn{1}{l}{}                              &     &      &      &           &        &        &      &       &        &     &        
\end{tabular}
}
\vspace{-1cm}
\label{tbl:syncompare}
\end{center}
\end{table*}
}

\begin{document}
\sloppy
\title{SPAC-Net: Synthetic Pose-aware Animal ControlNet for Enhanced Pose Estimation}


\author{Le Jiang    \and
        Sarah Ostadabbas 
}


\institute{L. Jiang and S. Ostadabbas \at
              Augmented Cognition Lab (ACLab), Electrical and Computer Engineering Department, Northeastern University, Boston, MA.
}

\date{Received: date / Accepted: date}

\maketitle

\begin{abstract}
Animal pose estimation has become a crucial area of research, but the scarcity of annotated data is a significant challenge in developing accurate models. Synthetic data has emerged as a promising alternative, but it frequently exhibits domain discrepancies with real data. Style transfer algorithms have been proposed to address this issue, but they suffer from insufficient spatial correspondence, leading to the loss of label information. In this work, we present a new approach called Synthetic Pose-aware Animal ControlNet (SPAC-Net), which incorporates ControlNet into the previously proposed Prior-Aware Synthetic animal data generation (PASyn) pipeline. We leverage the plausible pose data generated by the Variational Auto-Encoder (VAE)-based data generation pipeline as input for the ControlNet Holistically-nested Edge Detection (HED) boundary task model to generate synthetic data with pose labels that are closer to real data, making it possible to train a high-precision pose estimation network without the need for real data. In addition, we propose the Bi-ControlNet structure to separately detect the HED boundary of animals and backgrounds, improving the precision and stability of the generated data. Using the SPAC-Net pipeline, we generate synthetic zebra and rhino images and test them on the AP10K real dataset, demonstrating superior performance compared to using only real images or synthetic data generated by other methods. Our work demonstrates the potential for synthetic data to overcome the challenge of limited annotated data in animal pose estimation \footnote{The  code and synthetic datasets are available at \href{https://github.com/ostadabbas/SPAC-Net-Synthetic-Pose-aware-Animal-ControlNet}{https://github.com/ostadabbas/SPAC-Net-Synthetic-Pose-aware-Animal-ControlNet}}.

\end{abstract}

\synzebra
\section{Introduction}
Animal pose estimation has become a vital area of research in recent years, covering animal 3D model reconstruction \cite{zuffi2019three,biggs2020left,ruegg2022barc,li2021hsmal}, multi-animal pose estimation \cite{pereira2022sleap,marks2022deep}, and animal behavior analysis \cite{sheppard2022stride,ye2022panoptic}. However, the scarcity of annotated data is a significant challenge for developing accurate models in this domain. Synthetic data is a promising alternative due to its potential for cost-efficiency, customization, and the ability to simulate various environmental conditions \cite{zuffi20173d,varol2017learning,huang2021invariant}. Current methodologies for generating synthetic animal datasets involve 3D mesh models, lighting modulation, and superimposing rendered animal images onto arbitrary backgrounds \cite{zuffi2019three,mu2020learning,li2021synthetic}. However, synthetic data frequently exhibits domain discrepancies with real data, making it challenging to train models exclusively on synthetic data \cite{jiang2022prior}.

Style transfer algorithms have been proposed to address the synthetic vs. real domain gap by endowing synthetic data with realistic textures and lighting from real images \cite{nicegan,park2020contrastive,park2020swapping,kim2020deformable,kolkin2019style}. These methods have surprisingly good performance on style transformation tasks where spatial correspondence is not strictly required, such as landscape, and where image content is relatively simple, like animal faces. However, full-body animal style transformation requires the model to learn more precise spatial correspondences. Otherwise, this could result in the absence of body parts, such as the head or legs, during the style transfer process. In our previous work \cite{jiang2022prior}, we attempted to transfer the style of random natural scenes onto animal textures to increase the richness of animal textures and reduce the domain discrepancy with real data. Nonetheless, to achieve the high performance we had to mix in real images during the pose estimation model training process.

To overcome these limitations, we present a new approach called Synthetic Pose-aware Animal ControlNet for Enhanced Pose Estimation Network (SPAC-Net). We incorporated ControlNet \cite{control} into the previously proposed Prior-Aware Synthetic animal data generation (PASyn) pipeline \cite{jiang2022prior}, leveraging the plausible pose data generated by the Variational Auto-Encoder (VAE)-based synthetic animal data generation pipeline as input for the ControlNet Holistically-nested Edge Detection (HED) boundary \cite{xie2015holistically} task model. This approach allows us to generate synthetic data with pose labels that are closer to real data, making it possible to train a high-precision pose estimation network without the need for real data. In addition, we propose the Bi-ControlNet structure to separately detect the HED boundary of animals and backgrounds, improving the precision and stability of the generated data.

Using the SPAC-Net pipeline, we generate 3,000 synthetic zebra images and 2,400 synthetic rhino images and test them on the AP10K real dataset \cite{ap10k}. See some samples of synthetic animal images generated by SPAC-Net compared to other methods in \figref{synzebra}. The accuracy of the model trained exclusively on synthetic data outperforms the model trained on only 99 real images, and the accuracy of the model trained on both synthetic data and 99 real images achieves state-of-the-art performance. The stable diffusion model as part of the ControlNet offers a more robust alternative to traditional style transfer algorithms, with exceptional generalization capabilities and the potential to harness extensive training datasets. Overall, our work represents a significant contribution to the field of animal pose estimation, demonstrating the potential for synthetic data to overcome the challenge of limited annotated data in animal pose estimation.

\section{Related work}
It has been almost ten years since the early animal pose estimation work introduced. Yet, the performance of these models is still far inferior to the human pose estimation in terms of accuracy, cross-domain adaptation, and model robustness. This is mainly due to the lack of real-world data, which is the challenge for almost all animal pose estimation work. Numerable variety of species and subspecies, and considerable differences in physical characteristics and behavior patterns between them cause it difficult to form a labeled dataset with adequate samples. 

\subsection{Animal Pose Estimation with Synthetic Data}
Animal Pose Estimation with Synthetic Data Synthetic data is a promising substitute for real data in previous works. Coarse prior can be learnt and then it would be used to build pseudo-labels for enormous amounts of unlabeled real animal data \cite{li2021synthetic,mu2020learning}. The fly in the ointment is that works such as  \cite{li2021synthetic,mu2020learning} still use significant amounts of real data (such as TigDog dataset \cite{TigDOg}) in training, which may not be possible to access for the unseen species. The work in \cite{zuffi20173d}, which focuses on animal model recovery, also gives an extraordinary hint on this problem. It purposed a general 3D animal model (called SMAL) by learning the animal toys’ scan, and use the SMAL model and a few real pictures to generate a large amount of realistic synthetic data by adjusting the model’s texture, shape, pose, background, camera, and other parameters. They also trained an end-to-end network \cite{zuffi2019three} using only synthetic Grevy’s zebra data, which came from direct reconstruction from animal 2D images. However, their results are much worse than the current state-of-the-art in terms of 2D animal pose estimation. 

\subsection{Domain Gap between Real and Synthetic Data}
Synthetic data offers significant advantages in terms of cost and scalability; however, it suffers from a substantial domain gap when compared to real data, which will greatly reduce the improvement achieved through finetuning the model using synthetic data \cite{huang2021infant, jiang2022prior}. A primary cause of the domain gap is the lack of diverse and realistic textures, appearances, and poses for the synthetic animals. The synthetic data often relies on rendering a limited number of computer-aided design (CAD) models of animals. Consequently, the resulting synthetic animals cannot encompass the comprehensive set of variations necessary for training robust deep learning models. This limitation could lead to overfitting and poor generalization when models are applied to real data. Additionally, synthetic data generation lacks proper pose guidance due to the absence of large-scale pose datasets for animals, as seen with humans \cite{vposer}. The morphological differences among animals make it difficult to create a general generative model to design reasonable poses for synthetic animals, and most synthetic datasets end up with randomly generated poses. Another crucial source of the domain gap
arises from the incongruity between synthetic animals and real backgrounds. Directly pasting synthetic animals onto real backgrounds often results in strong inconsistencies in terms of projection, brightness, contrast, and saturation. Moreover, the synthetic data may not adequately capture the complex interplay of light, shadow, and texture that is commonly found in real-world scenarios, further exacerbating the domain gap. 
\PASyC

\subsection{Bridging the Domain Gap with Style Transfer}
Recent research has explored various techniques for mitigating the domain gap between real and synthetic data, such as domain adaptation, domain randomization, and adversarial training.  Some transformer-based \cite{deng2022stytr2} or CNNs-based \cite{li2017universal} style transfer methods can effectively achieve global style transfer, such as art genres and colors space. Some methods for image-to-image translation, such as Contrastive Unpaired Translation \cite{park2020contrastive} and Swapping Autoencoder \cite{park2020swapping}, which are based on StyleGAN  \cite{karras2020analyzing} have surprisingly good results in tasks where there is no need for precise spatial correspondence or where the images have a simple spatial relationship, such as landscape images, and human or animal faces. However, these methods are difficult to apply to style transfer tasks involving animals with diverse postures and complex structures, and there is a continuous need for novel strategies to enhance the utility of synthetic data in computer vision tasks. In recent years, both GAN-based and transformer-based style transfer methods become less appealing with the emergence of diffusion models \cite{sohl2015deep, ho2020denoising}. As likelihood-based models, diffusion models do not exhibit mode-collapse and training instabilities typically associated with GANs. Moreover, by heavily leveraging parameter sharing and multiple steps of the diffusion process, these models are capable of representing highly complex distributions of natural images \cite{rombach2022high}. The stable diffusion models \cite{rombach2022high} even exhibit superior capabilities in high-resolution image synthesis and diverse task performance by significantly reducing computational costs and ensuring highly accurate reconstructions with minimal latent space regularization. ControlNet \cite{control} takes full advantage of the benefits of stable diffusion models, providing a unified framework to process different task-specific conditions, including segmentation, depth map, canny edge, HED boundary map, etc. Consequently, we adopted ControlNet to replace the original transformer-based model, which significantly increased the diversity of the data and reduced the domain discrepancy between synthetic and real data.

\section{SPAC-Net: Synthetic Prior-aware Animal ControlNet}
Our novel synthetic prior-aware animal ControlNet (SPAC-Net) pipeline offers robust animal pose estimation, even under conditions of limited data availability and label discrepancies. The architecture of SPAC-Net is illustrated in \figref{PASyC}, and we leverage an off-the-shelf graphic engine (i.e., Blender \cite{Blender}) to render the 3D animal mesh into synthetic images. The SPAC-Net pipeline comprises three key components: (1) pose data augmentation, (2) synthetic data stylization, and (3) synthetic animal pose image dataset generation. In this work, we demonstrate the effectiveness of SPAC-Net using the zebra and rhino as our target animal for pose estimation.

\subsection{Capturing Animal Pose Prior}
\armature
When the real data of the target animal is insufficient, the prior knowledge of their pose can be learned from an ingeniously designed synthetic dataset. To learn the animal pose priors, we refer to the variational autoencoder (VAE) framework \cite{pavlakos2019expressive}. VAE has already been proven to be conducive to human pose priors learning, when it is trained on large-scale 3D human pose datasets \cite{vposer}. However, appropriate 3D datasets cannot be easily obtained for animal pose studies. Therefore, we feed several animated low-poly 3D animal models (inexpensively purchased from the CGTrader  \cite{Cgtrader}) to the VAE to learn the probabilistic distribution of the feasible poses. Then, the trained VAE serves as a generative model to create thousands of new poses which are used to rig the template animal meshes in Blender. As seen in \figref{armature}(a), the quadrupeds like horses and dogs are similar in their skeleton. Due to the similarity across quadrupeds, the structure \figref{armature}(b) is applicable to learning general animal pose priors. Also, considering legs’ considerable flexibility, we only focus on learning the priors of the legs in this work to simplify the model. In order to minimize the influence of skeleton sizes or bone lengths on training, angles between neighboring bones (12 of them) shown in \figref{armature}(c) are chosen as primary data for training animal leg pose priors.

\VAE
Our network follows the basic VAE process, including input, encoder, random sampling, decoder, and output as demonstrated in \figref{VAE}.  $A\in\mathbb{R}^{36}$ is a $ 1 \times 36 $ vector containing the angles between 12 pairs of adjacent bones, and each space angle is decomposed into three directions, and $\hat{A}$ is a vector of the same shape representing the generated angles as the model output. The role of the encoder which is composed of dense layers with rectified linear unit (ReLU)  activations is to calculate the mean $\mu$ and variance $\sigma$ of the network input. We use the reparameterization trick to randomly sample $z$ in latent space $Z\in\mathbb{R}^{16}$, where $z = \mu + \epsilon \times \sigma$ and $e$ satisfies the distribution in the shape of $\mathcal{N}(0,I)$. Finally, the decoder which has a mirror structure of encoder is used to reconstruct the set of space angle, and make the reconstruction results $\hat{A}$ as close to $A$ as possible. 

Considering that one infeasible angle can ruin an entire pose, during the pre-test, we first assign random sampling from a normal distribution to the variable $z$, and then we create the histograms of the 12 joint angles with enough $\hat{A}$ ($\approx$10,000 poses). Based on the statistical results, we can establish the sampling range for each angle. Each range is designed to remove the small amount of angles, which are far from the mean value, and all of the ranges are recorded. After the pre-test, when we generate the new poses, a `pose filter' serves to remove the pose $\hat{A}$ as long as one of the 12 angles is outside these specified ranges and finally produce the refined poses $\hat{A'}$.  We can also obtain poses with higher diversity by increasing the variance of the sampling distribution after determining the pose filter. The training loss of the VAE is:
\begin{align}
\label{eqn:kl}
    &\mathcal{L}_{\text {total}}=w_{1} \mathcal{L}_{KL}+w_{2} \nonumber \mathcal{L}_{\text {rec}},\\
    \mathcal{L}_{KL}=KL(q(z|A)||&  \mathcal{N}(0,I)), \,\,\, \text{and} \,\,\,  \mathcal{L}_{\text{rec}}=\|A-\hat{A}\|_{2}^{2},
\end{align}
where $w_{i}$ is the weight of each loss term. The Kullback-Leibler term, $\mathcal{L}_{KL}$, represent the divergence between the encoder’s distribution $q(z|A)$ and $\mathcal{N}(0,I))$.  $\mathcal{L}_{\text {rec}}$ is the reconstruction term. $\mathcal{L}_{KL}$ encourages a normal distribution noise while $\mathcal{L}_{\text{rec}}$, in contrast, encourages to reconstruct the $A$ without any divergence.

\subsection{Stylization: Blending into the Background}
 In our work, we utilized ControlNet \cite{control} as our primary tool for style transfer, leveraging the robust capabilities of Stable Diffusion models in high-resolution image synthesis. ControlNet's end-to-end architecture strikes a balance between retaining the broad capabilities derived from large-scale image training and adapting to the unique requirements of specific tasks. Crucially, its various conditional inputs, particularly edge map and segmentation map, paves the way for stylizing our synthetic animal images with pose labels. This significantly enhances the reduction of the domain gap between real-world and synthetic data.
\begin{align}
\label{eqn:control}
    L = \mathbb{E}_{z_0,t,c_t,c_f,\epsilon\sim N(0,1)}[\|\epsilon - \epsilon_{\theta}(z_t, t, c_t, c_f)\|^2_2]
\end{align}
where $z_0$ represents the input data encoded into the latent space by the SD model \cite{rombach2022high}. Diffusion process introduce noise to this latent representation to generate a progressively noisier image $z_t$, where $t$ signifies the number of times noise is applied. As $t$ increases, the image reaches a state resembling pure noise. $c_t$ and $c_f$ represent text prompts and task-specific conditions respectively. These, along with the time step $t$, guide the image diffusion algorithms to train an encoder denoted as $e_{\theta}$. $L$ stands for the comprehensive learning objective of the entire diffusion model, and can be directly applied when fine-tuning diffusion models \cite{control}. Here, we use the HED boundary map as the conditional input, as it does not generate overly complex intermediate results like the Canny edge, nor does it result in the loss of animal texture information, such as the zebra's stripes, like the segmentation map does. Therefore, the HED boundary emerges as an appropriate middle ground choice.

\textbf{Issues with Direct ControlNet Use:}
Although the method of integrating synthesized animal images and ControlNet provides us with a potential solution for generating diverse and realistic animal images, there are several issues associated with its direct use.  First, we attempt to directly use the synthesized animal data without background as the input for ControlNet. However, despite specifying environmental attributes such as 'mountain', 'grassland', and 'sunny', the lack of background texture in the HED boundary map leads to a rather monotonous outcome for the background. To mitigate this, we introduce a natural landscape dataset containing diverse environments, such as grasslands, savannas, farms, and forests, to enrich the background.

Nonetheless, another issue, which is shown in \figref{issuecontrolnet} surfaces. When the synthesized animal images are superimposed on these complex backgrounds, certain areas of the animal's body become indistinguishable due to overlapping boundaries with the background. As a result, when such a boundary map is processed by the stable diffusion model, it does introduce variety into the background. However, it also causes certain parts of the animal, particularly its limbs, to appear incomplete, blurred, and distorted. This indicates a necessity for further improvement in handling complex animal and background interactions within this approach.
\issuecontrolnet

\textbf{Our Version:} To overcome these issues, our proposed Bi-ControlNet structure, as shown in \figref{bicontrolnet}, uses two distinct HED boundary detectors to identify the boundaries of animals and backgrounds. This unique approach prevents complex background contours from interfering with the detection of zebra contours, which is often the case with standard ControlNet architectures. We process the animal and background boundaries separately and then overlay the animal boundary onto the background boundary. In the merged boundary map, the animal's boundary is fully preserved, and the background boundaries that fall outside the animal's boundary after the boundary maps overlap are also retained. This approach greatly enhances the accuracy of the final generated images, preventing the misidentification of animal body parts and ensuring a more plausible pose.


\bicontrolnet
\subsection{SPAC-Animals: A Synthetic Pose-aware Animal Dataset}
The SPAC-Animals dataset comprises 3,000 synthetic zebra images created by our synthetic prior-aware animal ControlNet (SPAC-Net) pipeline. To generate this dataset, we used zebra toys and textured zebra 3D models to synthesize the images with Blender. We employed the 3D synthetic model generation pipeline (3D-SMG) to reconstruct two models from zebra toys and adapted the remaining three from 3D models. Additionally, we created 3,000 synthetic rhino images using 5 artificial 3D meshes. We purchased one textured 3D model for each animal from CGTrader and rendered 600 images for each species. The animal pose in each image was generated by the VAE model, and we collected 300 real scenes of grass, savanna, and forest from the Internet to stylize the synthetic animal. The bone vector transformations were used to automatically generate annotations with each image by calculating the coordinates of each zebra joint in the pose.

\section{Experimental Analysis}
Firstly, we delineate the implementation specifics of SPAC-Net, including the selection criteria for the pose filter ranges. With the aim of attaining high prediction accuracy on "unseen" animals, given the availability of only a minimal quantity of real labeled data, we elect zebras and rhinos as the primary subjects to validate the universal efficacy of our SPAC-Net. The results were quantified using the PCK@0.05 metric. Furthermore, we conducted an ablation study to assess the individual contributions of each component of SPAC-Net, as well as to analyze the effects of various methods for synthetic data generation. 
\subsection{Implementation Details}

 \textbf{SPAC-Net:} We keep the setting in VPoser \cite{vposer} for training the VAE model. The learning rate of Adam optimizer is 0.001 and we set $w_{1}$ and $w_{2}$ which are the weights of $\mathcal{L}_{KL}$ and $\mathcal{L}_{\text{rec}}$ in \eqnref{kl}, as 0.005 and 0.01, respectively. The model is trained on 1,000 poses of animated realistic 3D models for 250 epochs with 128 poses in a batch. The training poses are extracted from the animated horse, dog, sheep, cow, and deer 3D models. Each of these animals has over 20 common actions designed by 3D animators, such as running, walking, and jumping. To increase the diversity of the poses, we choose to generate random samples from a Gaussian distribution $\mathcal{N}(0,2I)$. The sampling value range of the pose filter will be described in the \ref{posefilter}. For our domain transfer part, we use the pre-trained stable diffusion model provided by \cite{control}\\
 \label{posefilter}
 
\textbf{Pose Filter:} Based on the \figref{posefilter}, which is made by decoding of 10,000 random samples from a normal Gaussian distribution, we can set a special sampling range for each joint. The angle value within this range can be regarded as a valid angle, and the pose that satisfies the twelve ranges can be seen as an appropriate pose. These ranges are shoulder\_right [40, 100], elbow\_right [-125, 0], front-paw\_right [-25, 100], shoulder\_left [40, 100], elbow\_left [-125, 0], front-paw\_left [-25, 100], hip\_right [-120, -60], knee\_right [0, 80], back-paw\_right [-125, 0], hip\_left [-120, -60], knee\_left [0, 80], back-paw\_left [-125, 0]. In order to increase the variety of poses and generate more poses with angles near the boundary, we choose to generate random samples from a Gaussian distribution $\mathcal{N}(0,2I)$ after setting the pose filter. The dropout rate of the pose is 68.0\%.
\posefilter
\subsection{Evaluation Datasets}

\textbf{Zebra-300 Dataset} is the primary test set we use. It contains 40 images from the AP10K test set, 160 unlabeled images randomly selected from AP10K, and 100 images from the Grevy's zebra dataset \cite{zuffi2019three}. We labeled the images of the latter two according to the AP10K dataset. The images in Zebra-300 dataset are mostly taken in the wild, so the environment occlusion makes the pose estimation task challenging. 

\textbf{Zoo Zebra Dataset} has been compiled using pictures and videos captured at a local zoo. It includes 100 images, featuring 2 distinct individual mountain zebras. We performed manual animal pose labeling on the cropped and resized images in accordance with the label setting of the AP10K dataset. Each label incorporates 17 key points: nose, eyes, neck, shoulders, elbows, front paws, hips, knees, back paws, and the root of tail. The occlusion caused by the enclosure's fencing adds an extra layer of complexity to this dataset. The dataset has already been available online \cite{jiang2022prior}.

\textbf{Rhino-300 Dataset} serves as another crucial test set in our study. It comprises 40 images extracted from the AP10K test set and an additional 260 unlabeled images selectively drawn from the AP-60K dataset. The entirety of this dataset has been meticulously labeled by hand, following the annotation style of the AP10K dataset. The images within the Rhino-300 dataset primarily depict rhinos in natural habitats, presenting substantial environmental occlusion, which poses a significant challenge for pose estimation tasks. We plan to make this annotation publicly available alongside the publication of this paper.

\subsection{Pose Estimation Results}
\mainresult
In the experimental analysis, we denote the total number of real (R) images for targeted animals, real images excluding the targeted animal (AP10K), and synthetic images (SynAP and SPAC-Animals) used in model training with figures enclosed in brackets in \tabref{mainresult}. The datasets of SynAP and SPAC-Animals are partitioned into training and validation sets in a 7:1 ratio, following the AP10K setting. Due to the relative scarcity of real data, we adopt a more conservative split ratio of 4:1.

Examining  \tabref{mainresult}, it is observed that the zebra test set includes images not only from AP10K, but also from the more challenging Grevy's zebra dataset, which poses a greater difficulty for pose estimation. Consequently, when the model is trained with a mere 99 real images, the predictive accuracy for the rhino at 88.3\% appears higher than that of the zebra at 78.7\%. Training with only 99 images proves to be inadequate for either animal. Upon inclusion of all data (excluding the test animal) into the training process, a significant accuracy enhancement is observed for both species. This improvement is attributed to the transfer learning from data of different species that share similar structural characteristics \cite{cao2019cross}. Therefore, this experimental setup can be viewed as a baseline.

The incorporation of SynAP into the training set results in a dramatic increase in predictive results, equating to the baseline level. Interestingly, the model even surpasses the baseline for the zebra, achieving 92.4\% accuracy in contrast to the baseline's 91.4\%. 

Similarly, the addition of SPAC-Animals to the training set also notably augments model performance. For the rhino, while it was anticipated to surpass both the SynAP and the baseline results, it only managed to achieve parity with the SynAP outcomes.The primary reasoning for this, we hypothesize, is the rhino's lack of distinctive texture patterns for each body part. This often leads the ControlNet into erroneous judgements, such as mistaking the face for the hip. However, for animals exhibiting clear stripe patterns like zebras, SPAC-Net exhibits a remarkable performance boost, reaching an impressive 96.3\% accuracy.
\subsection{Ablation Study}
\abl
As observed in  \tabref{mainresult}, the model, even when trained on a meager amount of real data, is capable of accurately predicting keypoints with prominent features and limited flexibility, such as the nose and eyes. The prediction results, however, can be significantly influenced by the large degrees of freedom and multiple alternative candidate positions associated with limb keypoints. The primary enhancement introduced by SPAC-Net is the considerable augmentation of limb keypoint prediction. Also, we observe that for the zebra, the limbs, including the front paw, back paw, front knee, and back knee, are more challenging to achieve high accuracy compared to the rhino. This is due to the zebra's longer and thinner legs, which have higher flexibility, making them more difficult to recognize. Therefore, we focus our analysis on zebras here. To validate that the VAE, style transfer can alleviate the mismatching limb prediction and diminish the domain discrepancy between the real (R) and synthetic (S) domains, we conduct tests on the Zoo Zebra and Zebra-300 datasets.

 \tabref{abl} presents the results of eight experiments (a-h) conducted under varied training conditions. In the absence of VAE during training, we establish the range based on the available animal pose/animation data, uniformly sample the pose data from this range, and execute random sampling, akin to the methodology employed in \cite{zuffi2019three,mu2020learning}. In cases where StyTr$^2$ is not utilized, we retain the original texture of the model. The comparative analysis of (b), (c), (d), and (e), (f), (g) clearly demonstrates that models trained with both VAE and StyTr$^2$ achieve noticeably superior accuracy compared to those trained without them. 
\subsection{Comparative Analysis of Synthetic Data Methods}
\syncompare
\textbf{Quantitative Analysis:} In this section, we validate the superiority of our proposed SPAC-Net and Bi-ControlNet architectures by training exclusively on synthetic images generated by various methods. These methods include simple synthetic (simply pasting synthesized animals onto random backgrounds without any style transfer), SynAP (synthetic images created using our prior work, PASyn \cite{jiang2022prior}, with style transfer performed by  StyTr$^2$), ControlNet (data generated directly by feeding SynAP data to ControlNet \figref{issuecontrolnet}), and Bi-ControlNet (data generated using SPAC-Net \figref{bicontrolnet}). We further compare these training results with those obtained from training solely on 99 real images.
\ablation
As illustrated in Figure \figref{ablation}, it becomes apparent that synthetic animal images generated by the standard ControlNet encounter difficulties in producing clear body parts in corresponding regions when confronted with complex backgrounds or challenging poses. Notably, accurately restoring the orientation of the animal's face and the posture of the legs in the input animal images in the final images proves to be challenging. By independently conducting HED boundary detection on the animals, the stable diffusion model can discern the animal contours more distinctly and make accurate judgments.

The results in Table \tabref{syncompare} further emphasize that the pose estimation model, when trained on data generated by Bi-ControlNet, can achieve markedly superior performance than those trained using the standard ControlNet or SynAP. For the zebra, the accuracy even surpasses that of the real 99, achieving the highest accuracy at 80.4\%. This highlights the effectiveness of our proposed Bi-ControlNet approach in managing complex backgrounds and challenging poses while ensuring high-quality animal synthesis and maintaining pose estimation accuracy.
\tsne

\textbf{Visual Analysis:} We employ t-distributed stochastic neighbor embedding (t-SNE) to visualize 750 images each from four different domains—real (red), simple synthetic (purple), SynAP (blue), and SPAC-animals (green)—to verify the domain shift happening after style transfer. In this work, we utilize the second stage features of the High Resolution Network (HRNet) for t-SNE visualization due to their distinct advantages. Stage-2 features capture a balanced blend of basic visual cues and higher-level semantic information. Crucially, they retain texture-related characteristics, enabling us to distinguish subtle style variations across different image domains effectively. This makes stage-2 features a suitable choice for our task over other stages in HRNet.

As seen in \figref{tsne}, despite SynAP demonstrating superior results in the quantitative tests compared to simple synthetic,  their domain distributions largely overlap, making them challenging to distinguish. In contrast, SPAC-Animals exhibits a distinct domain shift compared to simple synthetic and SynAP. Further, the compact and distinct distribution of real data and SPAC-Animals suggests a strong similarity in their feature characteristics, facilitating the mapping between these two domains. Despite the lack of overlap between the clusters of real data and SPAC-Animals, their proximity in the t-SNE space indicates a high degree of similarity between their respective domains. 
The similarities in feature distribution between them are advantageous for transfer learning, which is further corroborated by the fact that a model pre-trained on real data (e.g., AP10K) and fine-tuned on SPAC-Animals would likely exhibit better performance. Conversely, the dispersed distribution of SynAP and simple synthetic data in the t-SNE visualization reflect more diverse or disparate domains compared to real data, which aligns with its lower performance in pose estimation tests.

\section{Conclusion and Future Work}
Our proposed SPAC-Net pipeline has demonstrated the potential of synthetic data for animal pose estimation, particularly in scenarios where the availability of real data is limited. By leveraging the VAE-based synthetic animal data generation pipeline as input for the ControlNet HED Boundary task model, we were able to generate synthetic data that is much closer to real data, reducing the domain gap and enabling us to train a high-precision pose estimation network without relying on real data. However, there are still some limitations to our approach. While the Bi-ControlNet has significantly reduced the incorrect predictions of animal structure and posture by Stable Diffusion, there is still room for improvement. In particular, we aim to find iterative optimization strategies and fine-tune the ControlNet model to minimize the misjudgment of limb parts caused by the lack of specific texture patterns, such as rhinos. Another limitation is the computational cost of Denoising Diffusion Implicit Models (DDIM) sampling in ControlNet, which may hinder its integration into end-to-end pose estimation models. To overcome this limitation, we will investigate alternative sampling strategies to accelerate the image generation process. Finally, we will investigate the use of unsupervised learning techniques to further reduce the reliance on labeled data and enhance the generalization capability of the pose estimation model.


%
%

\bibliographystyle{spbasic}      
\bibliography{ref}   

\end{document}